\documentclass[]{article}    
\usepackage[backref]{hyperref}  
\usepackage{amsmath, amssymb, natbib, graphicx, url, algorithm2e, amsthm}
\usepackage{times}  

\usepackage[english]{babel}
\usepackage[utf8]{inputenc}

\usepackage{booktabs}
\usepackage{color}
\usepackage{subfig}

\usepackage{tcolorbox}

\usepackage{algorithmic}

\usepackage{floatrow,fr-subfig}
\usepackage{dsfont}  
\usepackage{framed}
\usepackage[margin=0.8in]{geometry}

\usepackage[makeroom]{cancel} 

\newcommand{\E}{\mathbb{E}}
\renewcommand{\P}{\mathbb{P}}
\newcommand{\N}{\mathbb{N}}

\newcommand{\bigo}{\mathcal{O}}

\newcommand{\ONE}{\mathds{1}}

\DeclareMathOperator{\argmax}{argmax}

\DeclareMathOperator{\perm}{perm}

\newtheorem{theorem}{Theorem}[section]

\author{ {\bf Stefano Trac\`a} \\
	MIT \\
	stet@alum.mit.edu\\
	\and
	{\bf Cynthia Rudin}  \\
	Duke University          \\
	cynthia@cs.duke.edu \\
	\and
	{\bf Weiyu Yan}   \\
	Duke University \\
	weiyu.yan@duke.edu    \\
}

\theoremstyle{remark}

\theoremstyle{example}

\theoremstyle{proposition}
\newtheorem{proposition}{Proposition}

\begin{document}
\date{}
\title{Reducing Exploration of Dying Arms in Mortal Bandits}
\maketitle

\begin{abstract}
Mortal bandits have proven to be extremely useful for providing news article recommendations, running automated online advertising campaigns, and for other applications where the set of available options changes over time. Previous work on this problem showed how to regulate exploration of new arms when they have recently appeared, but they do not adapt when the arms are about to disappear. Since in most applications we can determine either exactly or approximately when arms will disappear, we can leverage this information to improve performance: we should not be exploring arms that are about to disappear. We provide adaptations of algorithms, regret bounds, and experiments for this study, showing a clear benefit from regulating greed (exploration/exploitation) for arms that will soon disappear. We illustrate numerical performance on the Yahoo! Front Page Today Module User Click Log Dataset. 
\end{abstract}

\section{INTRODUCTION}


In many applications of multi-armed bandits, the bandits are \textit{mortal}, meaning that they do not exist for the full period over which the algorithm is running. In advertising, ads and coupons can come and go; in news article recommendation, the news is perpetually changing; in website optimization, the content changes to keep viewers interested. \citet{chakrabarti2009mortal} introduced and formalized the notion of mortal bandits, and there has been a body of work following this.  
This work has proved to be valuable in the setting of advertising (see \citet{agarwal2009explore} and  \citet{Feraud:ug}) and in other areas such as communications underlaying cellular networks (see \citet{maghsudi2015channel}). \citet{bnaya2013volatile} propose an adaptation to the mortal settings of the popular UCB algorithm introduced by \citet{auer2002finite}. While these algorithms are designed to adapt exploration based on when arms \textit{appear}, they do not adapt when arms \textit{disappear} (for example, in the work of \citet{bnaya2013volatile}, new arms are immediately played, even for arms that may soon die, which could be a poor strategy). In strategic implementations of mortal bandits, \textit{we should not be exploring arms that are soon going to disappear}.

In the applications discussed above (advertising, news article recommendation, website optimization) and others, we often know in advance when arms will appear or disappear. For coupons and discount sales, we launch them for known periods of time (e.g., a one day sale), whereas for news articles, we could choose to place them in a pool of possible featured articles for mobile devices for one day or one week. If the lifespans of the arms are not known, they can often be estimated. For instance, we can observe the distribution of the lifespans of the arms to determine when an arm is old relative to other arms. Alternatively, external features can be used to estimate the remaining lifespan of an arm.

This work provides algorithms for the mortal bandit setting that reduce exploration for dying arms. 
In Section \ref{Section::Algorithms} we introduce two algorithms: the AG-L algorithm (adaptive greedy with life regulation) and the UCB-L algorithm (UCB mortal with life regulation). We present finite time regret bounds (proofs are in the Supplement\footnote{The Supplement is available in the GitHub repository: \url{https://github.com/5tefan0/Supplement-to-Reducing-Exploration-of}\\\url{-Dying-Arms-in-Mortal-Bandits} }) and intuition on the meaning of the bounds.  In Section \ref{Section::Experiments} we discuss numerical performance on the publicly available Yahoo$!$ Front Page Today Module
User Click Log Dataset. The experiments show a clear benefit in final rewards when the algorithms reduce exploration of arms that are about to expire. This confirms the intuition that it is useless to gain information about arms if they are going to disappear soon anyway.    

\section{ALGORITHMS FOR REGULATING EXPLORATION OVER ARM LIFE}\label{Section::Algorithms}

Formally, the mortal stochastic multi-armed bandit problem is a game played in $n$ rounds. At each round $t$ the algorithm chooses an action $I_t$ among a finite set $M_t$ of  possible choices called \emph{arms} (for example, they could be ads shown on a website, recommended videos and articles, or prices). When arm $j \in M_t$ is played, a random reward $X_j(t)$ is drawn from an unknown distribution. The distribution of $X_j(t)$ does not change with time (the index $t$ is used to indicate in which turn the reward was drawn) and it is bounded in $[a,b]$ (and we denote with $r$ the range $r=b-a$), while the set $M_t$ can change: arms may become unavailable (they ``die'') or new arms may arrive (they ``are born''). At each turn, the player suffers a possible regret from not having played the best arm: the mean regret for having played arm $j$ at turn $t$ is given by $\Delta_{j,i^*_t}=\mu_{i^*_t}-\mu_j$, where $\mu_{i^*_t}$ is the mean reward of the best arm available at turn $t$ (indicated by $i^*_t$) and $\mu_j$ is the mean reward obtained when playing arm $j$. Let us call $I(j)$ the set of turns during which the algorithm chose arm $j$. At the end of each turn the algorithm updates the estimate of the mean reward of arm $j$:
	\begin{equation}\label{Equation::mean_estimator}
	\hat{X}_{j} = \frac{1}{T_j(t-1)}\sum_{s \in I(j)}^{T_j(t-1)} X_j(s),
	\end{equation}
	where $T_j(t-1)$ is the number of times arm $j$ has been played before round $t$ starts. 
	
	Let us define $M_t$ as the set of all available arms at turn $t$ ($M_1$ is the starting set of arms). $M_I = \{1, 2, \dots, m_I\} \subset M_1, M_I \neq \emptyset$ is the set of arms that are initialized over the first $m_I$ iterations (i.e., the algorithm plays one time all of them following the order of their index).
	The quantity that a policy tries to minimize is the cumulative regret $R_n$ that is given by
	\begin{equation}\label{Equation::cumulative_regret}
	R_n=\sum_{j \in M_I}\Delta_{j,i^*_{j}} +    \sum_{t=m_I + 1}^n \;\sum_{j \in M_t} \Delta_{j,i^*_{t}} \ONE_{\{t \in I(j)\}}, 
	\end{equation}   
where $\ONE_{\{t \in I(j)\}}$ is an indicator function equal to $1$ if arm $j$ is played at time $t$ (otherwise its value is $0$). 
The first summation in \eqref{Equation::cumulative_regret} is the regret that the algorithm suffers during the initialization phase when each arm in $M_I$ is pulled once yielding a regret of $\Delta_{j,i^*_{j}}$ (the arms in $M_I$ are played in order of their index and $i^*_{j}$ denotes the best arm available at that turn). For the rest of the game ($t \in \{m_I + 1 , \cdots, n\}$), the algorithm incurs $\Delta_{j,i^*_{t}}$ regret at time $t$ only when arm $j$ is available ($j \in M_t$) and it is pulled ($t \in I(j)$). Let us call $M = \bigcup_{t=1}^n M_t$ the set of all arms that appear during the game and $L_j = \{s_j, s_j+1, \cdots, l_j\}$ the set of turns that arm $j$ is available. Then, we can also write \eqref{Equation::cumulative_regret} as 
\begin{equation}\label{Equation::cumulative_regret_second_form}
R_n=\sum_{j \in M_I}\Delta_{j,i^*_{j}} +   \sum_{j \in M}\; \sum_{\substack{t \in L_j \\ t>m_I}}  \Delta_{j,i^*_{t}} \ONE_{\{t \in I(j)\}}.
\end{equation}
Depending on the algorithm used, one formulation may be more convenient than the other when computing a bound on the expected cumulative regret $\E[R_n]$. A complete list of the symbols used throughout the paper can be found in Supplement E.

\subsection{THE ADAPTIVE GREEDY WITH LIFE REGULATION (AG-L) ALGORITHM}

In Algorithm \ref{Algorithm::AG-L} we extend the adaptive greedy algorithm (which we abbreviate with AG) presented in \citet{chakrabarti2009mortal}. We call this new algorithm the \textit{adaptive greedy with life regulation algorithm}, which we abbreviate with AG-L. AG-L handles rewards bounded in $[a,b]$, and regulates exploration based on the remaining life of the arms (that is, the algorithm avoids exploring arms that are going to disappear soon).  During the initialization phase, the algorithm plays each arm in the initialization pool $M_I$ once. After that, to determine whether to explore arms, AG-L draws from a Bernoulli random variable with parameter $$ p = 1-\frac{\max_{j \in M_t}\hat{X}_{j} -a}{b-a}, $$ which intuitively means that if the algorithm has a good available arm (i.e., an arm that has a high mean estimate) the probability of exploration is very low and the algorithm will exploit by playing the best available arm so far (ignoring also arms that were excluded in the initialization phase or have been born and never played). If the value of the Bernoulli random variable is $1$, then AG-L proceeds by playing an arm at random among those arms whose remaining life is long enough: we call this set $M_t(\mathcal{L})$. One way to set $M_t(\mathcal{L})$ is to pick all arms in $M_t$ such that their remaining lifespan is in the top 30\% of the distribution of all remaining lifespans (we chose 30\% because we tuned this parameter by trying different values on a small subset of data). $M_t(\mathcal{L})$ can also contain arms that have never been played before or that were excluded in the initialization phase. As mentioned earlier, if the value of the Bernoulli random variable is $0$, then AG-L exploits the arm that has the highest average reward. 

Note that this algorithm is not relevant to sleeping bandits (see \citet{kleinberg2010regret} and \citet{kanade2009sleeping}) because those arms do not die, they simply sleep. For sleeping bandits, we would want to explore them until they fall asleep because the estimate of the arm's mean reward would still be useful when the arm wakes up again.


\RestyleAlgo{boxruled}
\begin{algorithm}[t]
	\caption{AG-L algorithm}\nllabel{Algorithm::AG-L}
	\SetKwInOut{Input}{Input}
	\SetKwInOut{Output}{output}
	\SetKwInOut{Loop}{Loop}
	\SetKwInOut{Initialization}{Initialization}
	\Input{number of rounds $n$, initialization set of arms $M_I$, set $M_t$ of available arms at time, rewards range $[a,b]$  }
	\Initialization{play all arms in $M_I$ once, and initialize $\hat{X}_{j}$ for each $j=1,\cdots,m_I$}
	\For{$t=m_I+1$ \KwTo $n$}{ 
		{Draw a Bernoulli $B$ r.v. with parameter $$ p = 1-\frac{\max_{j \in M_t}\hat{X}_{j} -a}{b-a};$$}
		\If{$B = 1$}{Play an arm at random from $M_t(\mathcal{L})$\;
			\Else{Play an arm $j$ with highest $\hat{X}_{j}$\;}
		}
		{Get reward $X_j(t)$\;}
		{Update $\hat{X}_{j}$\;}
	}
\end{algorithm}

In order to derive a finite time regret bound we introduce $\mathcal{H}_{t-1}$ as the set of all possible histories (after deterministic initialization) of the game up to turn $t-1$:
\begin{align}\label{Equation::H}
    \mathcal{H}_{t-1} = \left\{ 
    h = \begin{bmatrix}
            b_{m_I+1} & b_{m_I+2} & \hdots & b_{t-1} \\
            i_{m_I+1} & i_{m_I+2} & \hdots & i_{t-1}
        \end{bmatrix} \textrm{such that}\right.\nonumber\\
        \left. b_s \in \{0,1\} , \; i_s \in M_s, \;\;\,\forall s \in \{m_I+1 , \hdots, t-1\}\nonumber
    \right\}.
\end{align}
Each element $h$ of $\mathcal{H}_{t-1}$ is a possible history of pulls before turn $t$ and tells exactly what arm was pulled and if it was an exploration turn or an exploitation turn. 
If $b_s = 1$ we say that the algorithm explored at time $s$, if $b_s = 0$ we say that the algorithm exploited at time $s$, while $i_s$ is the index of the arm that was played at time $s$. Let us define the linear transformation $g(p) = b+(a-b)p$ (used to standardize rewards to the interval $[0,1]$) and use a result from \citet{Vaughan1972permanent} for the PDF (or PMF) $f_{M(h,s)}(g(p))$ of the maximum of the estimated mean rewards at time $s$ given that each arm has been pulled according to history $h$ up to time $s-1$:
\footnotesize
\begin{equation*}
f_{M(h,k)}(x)=
\frac{1}{(m_t-1)!} \perm\left(    
\begin{bmatrix}
F_{1}(x)       &   \dots & F_{m_k}(x)   \\
\vdots         &   \ddots& \vdots       \\
F_{1}(x)       &   \dots & F_{m_k}(x)   \\
f_{1}(x)       &   \dots & f_{m_k}(x)  
\end{bmatrix}
\right)  
\end{equation*}\normalsize
where the matrix has a total of $m_k$ rows (and columns), $f_{1}(x), \cdots, f_{m_k}(x)$ and $F_{1}(x), \cdots, F_{m_k}(x)$ are the PDFs (or PMFs) of the distributions of the average rewards (which we can compute knowing the distribution from which rewards are drawn). For each $h$, we indicate how many times arm $j$ has been pulled up to time $k$ with 
\[t_j(h,k)= \ONE_{\{j \in M_I\}} +  \sum_{s'=m_I+1}^k \ONE_{\{i_{s'} \in I(j)\}}. \] Similarly to when we defined the regret, let us call $\Delta(i,i_s) = \mu_i-\mu_{i_s}$. 
Then, consider the following quantities (see Supplement A 
for how to compute them given the mean rewards):  
\begin{itemize}
\item $u_s(h,i_s)$ \textbf{is an upper bound on the probability that arm $i_s$ is considered to be the best arm at time $s$ given the history of pulls} (according to $h$) up to time $s-1$:
\begin{equation*}
	u_s(h,i_s) = \prod_{i: \mu_i > \mu_{i_s}} \left(  \exp \left\{ -\frac{t_{i_{s}}(h,s) \Delta(i,i_s)^2}{2 r} \right\} +   \exp \left\{ -\frac{t_{i}(h,s) \Delta(i,i_s)^2}{2 r} \right\} \right),
\end{equation*}
where range of rewards $r$ is defined as $r=b-a$.
\item $U_k(h,i_k)$ \textbf{is an upper bound on the probability that arm $i_k$ would be pulled at time $k$ given the history of pulls} (according to $h$) up to time $k-1$:\\
When $k < t$, then $U_k(h,i_k) =$
\begin{eqnarray}\nonumber
	\int_{0}^{1} \left( \frac{p}{m_{k}} \ONE_{\{b_k = 1\}}  +(1 - p) u_k(h,i_k) \ONE_{\{b_k = 0\}} \right) f_{M(h,k)}(g(p))\;
	\text{d}p,  &\label{Equation::U__s}
\end{eqnarray}
and when $k=t$, then $U_t(h,i_t) =$
\begin{eqnarray}
	\int_{0}^{1} \left( \frac{p}{m_{t}} + (1 - p) u_t(h,i_t) \right) f_{M(h,t)}(g(p))\;
	\text{d}p.
\end{eqnarray}
\item $U_t(h,j)$ \textbf{is an upper bound on the probability that arm $j$ would be pulled at time $t$ given the history of pulls} (according to $h$) up to time $t-1$:
\begin{eqnarray}\label{Equation::U__t}
	U_t(h,j) = \int_{0}^{1} \left( \frac{p}{m_{t}}    +
	(1 - p) u_t(h,j) 
	\right) f_{M(h,t)}(g(p))\;\text{d}p.
\end{eqnarray}
\end{itemize}
In standard regret bounds, the bound is usually in terms of the mean rewards $\mu_j$ and $\Delta_j$ for each arm $j$, which are not known in the application. Our bounds analogously depend on the $\mu_j$'s and $\Delta(i,i_s)$'s (where $i_s$ is the arm played at time $s$, and $i$ is another arm with higher mean reward). While standard bounds usually have a simple dependence on $\mu_j$'s, our bounds have a more complicated dependence on the $\mu_j$'s. On the other hand, they depend on the same quantities as the standard bounds; once we have the $\mu_j$ terms, the bound can be computed using the same information that is available in the standard bounds. For instance $u_s(h,i_s)$, $U_s(h,i_s)$, and $U_t(h,j)$ do not require any additional information other than the $\mu_j$'s.

Theorem \ref{Theorem::AG-L} presents a finite time upper bound on the regret for the AG-L algorithm (Supplement A 
has the proof).
\tcbset{colback=blue!2!white}
\begin{tcolorbox}
\begin{theorem}\label{Theorem::AG-L}
	The bound on the mean regret $\E[R_n]$ at time $n$ is given by
\begin{align}\label{Equation::Theorem::AG-L::1}
	&\E[R_n] 
	 \leq \sum_{j \in M_I} \Delta_{j,i^*_{j}} + \sum_{t=m_I + 1}^n \;\sum_{j \in M_t(\mathcal{L})} \Delta_{j,i^*_{t}} 
	 \sum_{h \in \mathcal{H}_{t-1}} \left(U_t(h,j)  \prod_{s = m_I+1}^{t-1} U_s(h,i_s)\right).
\end{align} 
\end{theorem}
\end{tcolorbox}

The standard case, when there is no exploration regulation based on remaining arms life, can be recovered by setting $M_t(\mathcal{L}) = M_t$. (This is the case where we are not excluding arms that are about to disappear). In that standard case, Theorem \ref{Theorem::AG-L} is a novel finite time regret bound for the standard AG algorithm introduced by \citet{chakrabarti2009mortal}.  

The first summation in \eqref{Equation::Theorem::AG-L::1} represents the total mean regret suffered during the initialization phase. Intuitively, it is the summation of the mean regrets $\Delta_{j,i^*_{t}}$ for having pulled an arm $j$ that is in the initialization set $M_I = \{1, 2, \dots, m_I\}$. The second triple summation in \eqref{Equation::Theorem::AG-L::1} represents the total mean regret suffered after the initialization phase. Intuitively, it is the summation of all the mean regrets $\Delta_{j,i^*_{t}}$ for having pulled an arm $j$ weighted by the bound on the probability of pulling arm $j$. The bound on the probability of pulling arm $j$ is computed by considering all possible histories of pulls up to turn $t-1$ (hence the summation over $\mathcal{H}_{t-1}$). For each history $h$ in the sum, the bound of choosing arm $j$ at time $t$ is given by multiplying the bound $U_t(h,j)$ on the probability of pulling arm $j$ at turn $t$ given $h$ with the bound on the probability of that particular history $h$ (given by the product of $U_s(h,i_s)$ up to turn $t-1$). 

To intuitively see why this regret bound is better than the one that arises from the standard AG policy, we look at the quantities in Equation \eqref{Equation::U__s}. The integrand has two main terms that are mutually exclusive (i.e., one appears during exploration turns and the other during exploitation turns):
\begin{itemize}
\item $1/m_s$ (recall that $m_s$ is the number of arms available at turn $s$): this is a constant appearing during exploration phases (when $b_s=1$). 
\item $u_s(h,i_s)$: this is a product of negative exponentials that decreases quickly, becoming smaller than $1/m_s$ after enough pulls on arm $i_s$. It appears during exploitation turns (when $b_s=0$).
\end{itemize}
The two terms are mutually exclusive, and the AG algorithm that explores more often will have the term $1/m_s$ appear more often in the integrand of Equation \eqref{Equation::U__s}. A larger integrand will yield a larger regret bound.


Conversely, the AG-L algorithm considers only the set $M_t(\mathcal{L})$ of arms with long life, and the term $1/m_s$ will appear less often than the smaller quantity $u_s(h,i_s)$, yielding a smaller regret bound.

Algorithms with smaller regret bounds generally lead to smaller regrets in practice. We will show how this is realized in the experiments later.

We can see the bound's intuition by restating Theorem \ref{Theorem::AG-L} with dependence on the $1/m_s$ and $u_s(h,i_s)$ terms notated explicitly:
\tcbset{colback=blue!2!white}
\begin{tcolorbox}
	\begin{theorem}\label{Theorem::AG-L-bound}
		The bound on the mean regret $\E[R_n]$ at time $n$ is given by
		\begin{align*}
    		& \E[R_n]\leq \bigo(1) + \sum_{t=m_I + 1}^n \sum_{j \in M_t(\mathcal{L})} \Delta_{j,i^*_{t}} 
    		 \sum_{h \in \mathcal{H}_{t-1} } F\left(\frac{1}{m_1}, \cdots, \frac{1}{m_t},  u_1(h,i_1), \cdots, u_t(h,i_t)\right),
		\end{align*}
		where
		\begin{align*}
		    & F\left(\frac{1}{m_1}, \cdots, \frac{1}{m_t},  u_1(h,i_1), \cdots, u_t(h,i_t)\right) 
		     = U_t(h,j)  \prod_{s = m_I+1}^{t-1}	U_s(h,i_s)
		\end{align*}
		is an increasing function of all its arguments. 
	\end{theorem}
\end{tcolorbox}
Intuitively, $u_1(h,i_1), \cdots, u_t(h,i_t)$ are smaller than $\frac{1}{m_1}, \cdots, \frac{1}{m_t}$ since they decrease at a fast rate (they are products of negative exponentials). By regulating exploration on arms that live longer, the bound of Algorithm \ref{Algorithm::AG-L} presents the smaller terms more times than the larger ones, yielding an overall better expected regret. Reducing exploration on dying arms tends not to impact the other reward terms unless the dying arms have significantly better rewards than the long-lived arms, which generally is not the case in real applications.

\textbf{A thought experiment with good and bad arms.\\}
Let us conduct a thought experiment to provide intuition for why it is beneficial to limit exploration only among arms with short remaining life. Consider two different standard games, where arms are always available: the first with $100$ arms, and the second with $10$ arms. The quality of the arms come from the same distribution, for example we know that 30\% of the arms have high expected rewards, and 70\% have instead low expected rewards. The probability of picking a bad arm at random is the same in both games. However, one of these games is much more difficult than the other one in practice: in the 10-arm game, we can allocate more pulls to each arm, and thus it is much easier to determine when an arm is bad based on its mean reward estimate. For the 10-arm game, the algorithm will explore less (the term $1/m_s$ will appear less often) than in the 100-arm game, and thus the 10-arm game will have better bounds on the probability of playing suboptimal arms (the terms $u_s(h,i_s)$ decrease more quickly). Thus, it is easier to play the standard game with fewer arms.

When there is a mixture of long-lived and short-lived arms, AG-L may (in essence) reduce the full game to a smaller, easier one that considers only the long-lived arms.

In real applications, at each time, we expect there to be a mixture of arms with short remaining life and long remaining life. Intuitively, AG-L would reduce the game to an easier game by playing (among approximately good arms) mainly the long-lived arms.

\RestyleAlgo{boxruled}
\begin{algorithm}[]
	\caption{UCB-L algorithm}\nllabel{Algorithm::UCB-L}
	\SetKwInOut{Input}{Input}
	\SetKwInOut{Output}{output}
	\SetKwInOut{Loop}{Loop}
	\SetKwInOut{Initialization}{Initialization}
	\Input{number of rounds $n$, initialization set of arms $M_I$, set $M_t$ of available arms at time, rewards range $[a,b]$ }
	\Initialization{play all arms in $M_I$ once, and initialize $\hat{X}_{j}$ for each $j=1,\cdots,m_I$}
	\For{$t=m_I+1$ \KwTo $n$}{ 
		{Play arm with highest $\hat{X}_{j} + \psi(j,t)\sqrt{\frac{2 \log (t-s_j)}{T_{j}(t-1)}}$ \;}
		{Get reward $X_j(t)$\;}
		{Update $\hat{X}_{j}$\;}
	}
\end{algorithm}
\subsection{THE MORTAL UCB WITH LIFE REGULATION ALGORITHM (UCB-L)}
Algorithm \ref{Algorithm::UCB-L} extends the UCB algorithm of \cite{auer2002finite} to handle life regulation. In the standard UCB algorithm, the arm with the highest upper confidence bound above the estimated mean is played. In this new version, the upper confidence bound has been modified so that it can be used in the mortal setting. It gradually shrinks the estimated UCB as the life of the arm comes to an end. Exploration is thus encouraged only on arms that have a long lifespan. In this way, arms that are close to expiring are played only if their estimated mean is high. Let $s_j$ and $l_j$ be the first and last turn at which arm $j$ is available, and let $\psi(j,t)$ be a function proportional to the remaining life of arm $j$, which decreases over time. An example for $\psi(j,t)$ is $c\log(l_j - t +1)$, where $c$ is a positive constant (note that $\psi(j,t)$ approaches zero as the game gets closer to the expiration of arm $j$). New arms are initialized by using the average performance of past arms (i.e., if in the past, many bad arms appeared, new arms are considered more likely to be bad), and their upper confidence bound is built as if they have been played once. We abbreviate this algorithm by UCB-L.

Theorem \ref{Theorem::UCB-L} presents a finite time regret bound for the UCB-L algorithm (proof in Supplement B
).

\tcbset{colback=blue!2!white}
\begin{tcolorbox}
	\begin{theorem}\label{Theorem::UCB-L}
		Let $\bigcup_{z=1}^{E_j}L_j^z$ be a partition of $L_j$ into epochs with different best available arm, $s_j^z$ and $l_j^z$ be the first and last step of epoch $L_j^z$, and for each epoch let $u_{j,z}$ be defined as
		\begin{equation*}
		u_{j,z} = \max_{t\in\{s_j^z,\cdots,l_j^z\}}\left\lceil \frac{8 \psi(j,t) \log (t-s_j)}{\Delta_{j,z}^2}  \right\rceil,
		\end{equation*}
		where
		\begin{equation*}
		\Delta_{j,z} = \Delta_{j,i^*_{t}} \;\;\text{for}\;t \in L_j^z.
		\end{equation*}
		Then, the bound on the mean regret $\E[R_n]$ at time $n$ is given by
\begin{eqnarray*}
	\E[R_n]
	&\leq& \sum_{j \in M_I}\Delta_{j,i^*_{j}} +   \sum_{j \in M}\; \sum_{z =1}^{E_j}  \Delta_{j,z}\min\left(l_j^z-s_j^z \;,\; u_{j,z}
	\right.\\
	&&		+ \displaystyle\sum_{\substack{t \in L_j^z \\ t>m_I}}\; (t-s_{i^*_{t}}) (t-s_j-u_{j,z}+1)
	\left.\times\left[  (t-s_j)^{-\frac{4}{r^2}\psi(j,t)} +  (t-s_{i^*_{t}})^{-\frac{4}{r^2} \psi(i^*_{t},t)}  \right] \right) .
\end{eqnarray*}	\normalsize
	\end{theorem}
\end{tcolorbox}
The first summation $\sum_{j \in M_I}\Delta_{j,i^*_{j}}$ is the regret suffered during the initialization phase (the arms in $M_I$ are played in order of their index and $i^*_{j}$ denotes the best arm available at that turn). Intuitively, $u_{j,z}$ is the number of pulls required to be able to distinguish arm $j$ from the best arm in epoch $z$. In the second double summation, the mean regret for pulling arm $j$ is multiplied by the minimum between the epoch length and the upper bound on the probability that the arms \textit{appears} to be the best available one. This upper bound is a combination of the probability that we are either underestimating the best arm in epoch $z$ or we are overestimating arm $j$ (see Supplement B for more details).
If the game is such that no new arms are born during the game and all arms expire after turn $n$, then this regret bound reduces to the standard UCB bound
(see \cite{auer2002finite}).


\tcbset{colback=white}

\section{EXPERIMENTS ON Yahoo! NEWS ARTICLE RECOMMENDATION}\label{Section::Experiments}
We tested the performance the new AG-L and UCB-L algorithms versus the standard AG and UCB algorithms using the dataset from the Yahoo! Webscope program. The dataset consists of a stream of recommendation events that display articles randomly to users. At each time, the dataset contains information on the action taken (which is the article shown to the human viewing articles on Yahoo!), the outcome of that action (click or no click), the candidate arm pool at that time (the set of articles available) and the associated timestamp. We preprocessed the original text file into a structured data frame (see an extract of the data frame in Table \ref{dataframe}). 

\begin{table}
    \centering
    \caption{extracted dataframe from the original text record}
    \label{dataframe}
    \begin{tabular}{ | c | c | c | c |} 
        \hline
        timestamp & id & clicked & number of arms\\ 
        \hline
        1317513291 & id-560620 & 0 & 26 \\
        1317513291 & id-565648 & 0 & 26 \\
        1317513291 & id-563115 & 0 & 26 \\
        1317513292 & id-552077 & 0 & 26 \\
        1317513292 & id-564335 & 0 & 26 \\
        \hline
    \end{tabular}
\end{table}

In each game, the algorithms are tested on the same data. The recommender algorithms play for a fixed number of turns. We record the accumulated rewards of each algorithm.
At each time, a reward can be calculated only when the article that was displayed to the human user matches the action of the algorithm. (We do not know the outcome of actions not recorded in the dataset.) This means rewards can only be calculated at a fraction of times that the algorithm is playing. Therefore, while still playing the same number of turns, some algorithms will have more evaluations than others. In particular, an algorithm can be unlucky, in that most of its actions are discarded by chance. However, when the dataset was constructed, articles were shown uniformly at random to the human user, and overall,  the difference between the number of evaluations per algorithm is small. More details on the experiment can be found in Supplement D.

For ad serving or article serving in practice, the AG-L and UCB-L algorithms would be told when articles (or advertisements, or coupons) are scheduled to appear and expire. Accordingly, we provided the algorithms with the beginning and end of life for each arm.

Separately, we consider the case where we do not have the life of each arm in advance. In that case, we add a step to the algorithms, which estimates the lifespan of new arms by the mean lifespan of expired arms. 

\begin{figure}
    \centering
    \includegraphics[scale=0.6]{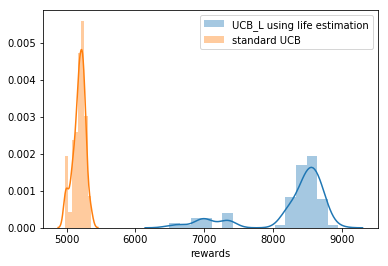}
    \caption{UCB's playing a set of 100 slightly different games. Randomness arises not from the algorithms but from random starting time.}
    \label{fig:UCBs}
\end{figure}

In order to obtain a distribution for performance (rather than a single performance measurement), we ran the AG and AG-L algorithms many times to plot the distribution of rewards. The AG and AG-L algorithms are non-deterministic, since they choose arms randomly from the candidates with enough remaining life. On the other hand, UCB and UCB-L algorithms  are deterministic because they always pick the arm with the best upper confidence bound. Running UCB and UCB-L many times on the same dataset will always give the same result. Therefore, to obtain a distribution for performance, we ran UCB and UCB-L for different sliding windows of time (i.e., we started the algorithms at many different points in time), which explains the multi-modal shape of UCB-L rewards distribution in Figure \ref{fig:UCBs}.

\begin{figure}
    \centering
    \includegraphics[scale=0.6]{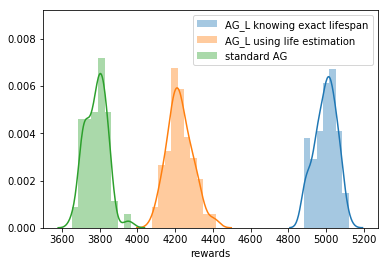}
    \caption{AG's playing the game 100 times. Randomness arises from the AG algorithms.}
    \label{fig:AGs}
\end{figure}

Figures \ref{fig:AGs} and \ref{fig:UCBs} show the empirical distribution of rewards for the algorithms. Each algorithm played 100 games with 100000 turns per game.  Each game consumed millions of data rows, because many actions could not be evaluated, as discussed above (they did not match the action shown to the Yahoo! user at that time). 

The algorithms with life-regulation dramatically outperform the standard ones. Among AG-Ls, knowing the exact lifespan of each article (rather than using an estimated lifespan) improves performance. This result would have been obvious in retrospect: more information given to the algorithm allows it to make better decisions.

The AG-L strategy adopted here was part of a high-scoring entry of one of the Exploration-Exploitation competitions. The entry scored second place, with a score that was not statistically significantly different from the first place entry. In this competition, AG-L was one of two key strategies contributing to the high score. Both key strategies were based on incorporating time series information about article behavior, which added more strategic value than other types of information available during the competition.

\section{CONCLUSIONS}
In this work, we have shown that it is possible to leverage knowledge about the lifetimes of the arms to improve the quality of exploration and exploitation in mortal multi-armed bandits. Our algorithms focus on exploring the arms that will be available longer, leading to substantially increased rewards. 
In cases where we do not know the lifetimes of the arms but can estimate them, these techniques are still able to substantially increase rewards.  
We have presented novel finite time regret bounds and numerical experiments on the publicly available Yahoo! Webscope Program Dataset that show the benefit of reducing exploration on arms that are about to disappear soon.

\bibliographystyle{plainnat}
\bibliography{mybibMortalMAB.bib}
\newpage
\newpage
\pagenumbering{gobble}
\appendix

\section{The regret bound of the Adaptive greedy algorithm}\label{Proof::AG-L_regret_bound}

We present a finite-time bound on the cumulative regret defined in Equation \eqref{Equation::cumulative_regret}.\\
Let $\mathcal{H}_{t-1}$ is the set of all possible histories (after deterministic initialization) of the game up to turn $t-1$:
\begin{equation}\label{Equation::H}
\mathcal{H}_{t-1} = \left\{ 
h = \begin{bmatrix}
b_{m_I+1} & b_{m_I+2} & \hdots & b_{t-1} \\
i_{m_I+1} & i_{m_I+2} & \hdots & i_{t-1}
\end{bmatrix}   
: b_s \in \{0,1\} , \; i_s \in M_s, \;\;\,\forall s \in \{m_I+1 , \hdots, t-1\}
\right\}.
\end{equation}
 If $b_s = 1$ we say that the algorithm explored at time $s$, if $b_s = 0$ we say that the algorithm exploited at time $s$, while $i_s$ is the index of the arm that was played at time $s$.
\tcbset{colback=blue!2!white}
\begin{tcolorbox}
	{\bf Theorem \ref{Theorem::AG-L}}\textit{
		Let us define the following quantities:
		\begin{itemize}
			\item $g(p) = b+(a-b)p$ ,
			\item $f_{M(h,s)}(g(p))$ is the PDF (or PMF) of the maximum of the estimated mean rewards at time $s$ given that each arm has been pulled according to history $h$ up to time $s-1$:
			\begin{equation*}
			f_{M(h,k)}(x)=
			\frac{1}{(m_t-1)!} \perm\left(    
			\begin{bmatrix}
			F_{1}(x)       & F_{2}(x) &  \dots & F_{m_k}(x)   \\
			\vdots         & \vdots   &  \vdots& \vdots       \\
			F_{1}(x)       & F_{2}(x) &  \dots & F_{m_k}(x)   \\
			f_{1}(x)       & f_{2}(x) &  \dots & f_{m_k}(x)  
			\end{bmatrix}
			\right)  
			\begin{array}{lc}
			\Bigg\}      & \vphantom{\rule{1mm}{27pt}} m_k -1 \;\text{rows}    \\
			\vphantom{\rule{1mm}{17pt}}         & \vphantom{\rule{1mm}{17pt}}   
			\end{array},
			\end{equation*}
			where $f_{1}(x), \cdots, f_{m_k}(x)$ and $F_{1}(x), \cdots, F_{m_k}(x)$ are the PDFs (or PMFs) of the distributions of the average rewards,
			\item $u_s(h,i_s)$ is an upper bound on the probability that arm $i_s$ is considered to be the best arm at time $s$ given the history of pulls (according to $h$) up to time $s-1$:
			\begin{equation*}
			u_s(h,i_s) = \prod_{i: \mu_i > \mu_{i_s}} \left(  \exp\left\{-\frac{t_{i_{s}}(h,s) \Delta(i,i_s)^2}{2 r}\right\} + \exp\left\{-\frac{t_{i}(h,s) \Delta(i,i_s)^2}{2 r}\right\}     \right), 
			\end{equation*}
			\item $U_s(h,i_s)$ is an upper bound on the probability that arm $i_s$ was pulled at time $s$ given the history of pulls (according to $h$) up to time $s-1$:
			\begin{equation*}
			U_s(h,i_s) = \int_{0}^{1} \left( \frac{p}{m_{s}}  \ONE_{\{b_s = 1\}}  +(1 - p) u_s(h,i_s) \ONE_{\{b_s = 0\}}
			\right) f_{M(h,s)}(g(p))\;\text{d}p,
			\end{equation*}
			\item $u_t(h,j)$ is an upper bound on the probability that arm $j$ is considered to be the best arm at time $t$ given the history of pulls (according to $h$) up to time $t-1$:
			\begin{equation*}
			u_t(h,j) = 
			\prod_{i: \mu_i > \mu_{j}} \left( \exp\left\{-\frac{t_{j}(h,t) \Delta(i,j)^2}{2 r}\right\} +  \exp\left\{-\frac{t_{i}(h,t) \Delta(i,j)^2}{2 r}\right\}\right),
			\end{equation*}
			\item $U_t(h,j)$ is an upper bound on the probability that arm $j$ was pulled at time $t$ given the history of pulls (according to $h$) up to time $t-1$:
			\begin{equation*}
			U_t(h,j) = \int_{0}^{1} \left( \frac{p}{m_{t}}    +
			(1 - p) u_t(h,j) 
			\right) f_{M(h,t)}(g(p))\;\text{d}p.
			\end{equation*}
		\end{itemize}
		Then, an upper bound on the expected cumulative regret $R_n$ at round $n$ is given by
		\begin{equation*}
		\E[R_n]\leq \sum_{j \in M_I}\Delta_{j,i^*_{j}} +    \sum_{t=m_I + 1}^n \;\sum_{j \in M_t} \Delta_{j,i^*_{t}} \sum_{h \in \mathcal{H}_{t-1}} \left(U_t(h,j)  \prod_{s = m_I+1}^{t-1}	U_s(h,i_s)\right).
		\end{equation*} 
}
\end{tcolorbox}
\tcbset{colback=white}
\begin{tcolorbox}
	{\bf First step:} Decomposition of $\E[R_n]$.
\end{tcolorbox}
\begin{equation}\label{Equation::ER_Adaptive_greedy}
\E[R_n]=\sum_{j \in M_I}\Delta_{j,i^*_{j}} +    \sum_{t=m_I + 1}^n \;\sum_{j \in M_t} \Delta_{j,i^*_{t}} \P\left(t \in I(j)\right), 
\end{equation}   
where we can write $\P\left(t \in I(j)\right)$ as
\begin{equation}\label{Equation::Decomposition}
\P\left(t \in I(j)\right) = \sum_{h \in \mathcal{H}_{t-1}} \P\left(t \in I(j)\; \Big|\; H_{t-1} = h \right)\P\left( H_{t-1} = h \right),
\end{equation}
where $H_{t-1}$ is a random variable that takes values in $\mathcal{H}_{t-1}$ defined as
\begin{equation}\label{Equation::H}
	\mathcal{H}_{t-1} = \left\{ 
	h = \begin{bmatrix}
	b_{m_I+1} & b_{m_I+2} & \hdots & b_{t-1} \\
	i_{m_I+1} & i_{m_I+2} & \hdots & i_{t-1}
	\end{bmatrix}   
	: b_s \in \{0,1\} , \; i_s \in M_s \;\;\,\forall s \in \{m_I+1 , \hdots, t-1\}
	\right\}.
\end{equation}
$\mathcal{H}_{t-1}$ is the set of all possible histories (after deterministic initialization) of the game up to turn $t-1$. If $b_s = 1$ we say that the algorithm explored at time $s$, if $b_s = 0$ we say that the algorithm exploited at time $s$, while $i_s$ is the index of the arm that was played at time $s$. The set $\mathcal{H}_{t-1}$ has $\prod_{s=m_{I}+1}^{t-1} (2m_s)$ elements. Note also that, by design of the algorithm, if an arm $j$ is new at time $s$, 
$$\P\left( H_{t-1} =   
\begin{bmatrix}
 b_{m_I+1} & \hdots &  b_s = 0 & \hdots &b_{t-1} \\
 i_{m_I+1} & \hdots &  i_{s} = j & \hdots & i_{t-1}
\end{bmatrix}
\right) = 0, $$ 
because the algorithm does not allow exploitation of a new arm. 
In the following steps we study (and find an upper bound when needed) each term in \eqref{Equation::Decomposition}.

\tcbset{colback=white}
\begin{tcolorbox}
	{\bf Second step:} Upper bound for $\P\left( H_{t-1} = h \right)$.
\end{tcolorbox}
Let us define $h_k$, with $k > m_I, \; k\in \N$, the first $k-m_I$ columns of $h$ (so $h$ and $h_{t-1}$ are the same).

For each $h$, we indicate how many times arm $j$ has been pulled up to time $k$ with 
\[t_j(h,k)= \ONE_{\{j \in M_I\}} +  \sum_{s=m_I+1}^k \ONE_{\{i_s \in I(j)\}},  \]
and, similarly to the definition of $\widehat{X}_j$ given in \eqref{Equation::mean_estimator}, we denote the mean estimated reward for arm $j$, given history of pulled arms $h$, with 
\begin{equation}\label{Equation::mean_estimator_deterministic_t}
\widehat{X}_{j}(h,k) = \frac{1}{t_j(h,k-1)}\sum_{s\in I(j)}^{t_j(h,k-1)} X_j(s).
\end{equation}

For each $h$, the probability of exploration at time $k$ is a random variable $E(h,k)$ with distribution given by 
\begin{equation}\label{Equation::probability_of_exploration0}
\P\left(E(h,k) = p \right) = \P\left(  1-\frac{\max_{j \in M_k}\widehat{X}_{j}(h,k) -a}{b-a} = p \right),
\end{equation}

Let us define $g(p) = b+(a-b)p$, then we can rewrite \eqref{Equation::probability_of_exploration0} as
\begin{equation}\label{Equation::probability_of_exploration}
\P\left(E(h,k) = p \right) = \P\left(\max_{j \in M_k}\widehat{X}_{j}(h,k) = g(p) \right).
\end{equation} 
We will give a formula for \ref{Equation::probability_of_exploration} in the next step of the proof.

 We can compute $\P\left( H_{t-1} = h \right)$ recursively using the fact that $\P\left( H_{t-1} = h \right)$ is equal to \footnotesize
\begin{equation}\label{Equation::H_t-1}
	\P\left( H_{t-1} = h \,\Big|\, H_{t-2} = h_{t-2}\right)\P\left( H_{t-2} = h_{t-2} \,\Big|\, H_{t-3} = h_{t-3} \right)\cdots\P\left( H_{m_I+2} = h_{m_I+2} \,\Big|\, H_{m_I+1} = h_{m_I+1} \right)\P\left(  H_{m_I+1} = h_{m_I+1} \right).
\end{equation}\normalsize
$h_{m_I+1}$ has only one column: $\begin{bmatrix} b_{m_{I} +1} \\ i_{m_I+1} \end{bmatrix}$, where $b_{m_{I} +1} \in \{0,1\}$ and $i_{m_I+1}\in M_{m_I+1}$.

We can write $\P\left(  H_{m_I+1} = h_{m_I+1} \right)$ as
\footnotesize
\begin{equation}
	\int_{0}^{1} \left( \frac{p}{m_{I+1}}  \ONE_{\{b_{m_I+1} = 1\}}  +
	(1 - p) \P\left(\widehat{X}_{i_{m_I+1}}(h,m_I+1) > \widehat{X}_{i}(h,m_I+1)  \;\;\forall i \neq i_{m_I+1}\right) \ONE_{\{b_{m_I+1} = 0\}}
	\right) \P\left(E(h,m_I+1) = p \right)\;\text{d}p
\end{equation}
\normalsize

Similarly, we can compute each term in \eqref{Equation::H_t-1}. For each $s \in \{m_I+2, \cdots, t-1\}$, we have that $\P\left( H_{s} = h \,\Big|\, H_{s-1} = h_{s-1}\right)$ is given by

\footnotesize
\begin{equation}
\int_{0}^{1} \left( \frac{p}{m_{s}}  \ONE_{\{b_s = 1\}}  +
(1 - p) \P\left(\widehat{X}_{i_{s}}(h,s) > \widehat{X}_{i}(h,s)  \;\;\forall i \neq i_{s}\right) \ONE_{\{b_s = 0\}}
\right) \P\left(E(h,s) = p \right)\;\text{d}p
\end{equation}
\normalsize

%

Using independence of the arms and Proposition \ref{Proposition::inclusion}, for each $s \in \{m_I+1, \cdots, t-1\}$ we can write  
\begin{eqnarray}
	&&\P\left(\widehat{X}_{i_{s}}(h,s) > \widehat{X}_{i}(h,s)  \;\;\forall i \neq i_{s}\right)\\ 
	&\leq& 
	\P\left(\widehat{X}_{i_{s}}(h,s) > \widehat{X}_{i}(h,s)  \;\;\forall i: \mu_i > \mu_{i_s}\right) \\
	&\leq&   \prod_{i: \mu_i > \mu_{i_s}} \P\left(\widehat{X}_{i_{s}}(h,s) > \widehat{X}_{i}(h,s)\right) \\
	&\leq& \prod_{i: \mu_i > \mu_{i_s}} \left[ \P\left(\widehat{X}_{i_{s}}(h,s) > \mu_{i_s} + \frac{\Delta(i,i_s)}{2}\right) +\P\left(\widehat{X}_{i}(h,s) < \mu_{i} - \frac{\Delta(i,i_s)}{2}\right)
	 \right]
\end{eqnarray}
and then bound each term by
 using Hoeffding's inequality\footnote{{\bf Hoeffding's bound:} Let $X_1, \cdots, X_n$ be r.v. bounded in $[a_i,b_i]$ $\forall i$. Let $\widehat{X} = \frac{1}{n}\sum_{i=1}^n X_i$ and $\mu = \E[\widehat{X}]$. \\Then, $\P\left(\widehat{X} - \mu \geq \varepsilon\right) \leq \exp\left\{ -\frac{2n^2\varepsilon^2}{\sum_{i=1}^n (b_i-a_i)^2 }  \right\}$. In our case, $\varepsilon = \frac{\Delta_{i_s,i}}{2}$, $n = t_s$ or $t_{i_s}$, $b-a = r$.}:
 \begin{equation}
 	\P\left(\widehat{X}_{i_{s}}(h,s) > 
 	\mu_{i_s} + \frac{\Delta(i,i_s)}{2}\right) 
 	\leq \exp\left\{-\frac{t_{i_{s}}(h,s) \Delta(i,i_s)^2}{2 r}\right\}
 \end{equation} 
 and
 \begin{equation}
  \P\left(\widehat{X}_{i}(h,s) < \mu_{i} - \frac{\Delta(i,i_s)}{2}\right) 
 \leq \exp\left\{-\frac{t_{i}(h,s) \Delta(i,i_s)^2}{2 r}\right\}.
 \end{equation}
 Let us define
 \begin{equation}
 	u_s(h,i_s) = \prod_{i: \mu_i > \mu_{i_s}} \left(  \exp\left\{-\frac{t_{i_{s}}(h,s) \Delta(i,i_s)^2}{2 r}\right\} + \exp\left\{-\frac{t_{i}(h,s) \Delta(i,i_s)^2}{2 r}\right\}     \right),
 \end{equation}
then, $ \P\left( H_{s} = h \,\Big|\, H_{s-1} = h_{s-1}\right) \leq U_s(h,i_s)$, where
\begin{equation}\label{Equation::U_s}
	U_s(h,i_s) = \int_{0}^{1} \left( \frac{p}{m_{s}}  \ONE_{\{b_s = 1\}}  +
	(1 - p) u_s(h,i_s) \ONE_{\{b_s = 0\}}
	\right) \P\left(E(h,s) = p \right)\;\text{d}p,
\end{equation}
and from \eqref{Equation::H_t-1}
\begin{equation}
	\P\left( H_{t-1} = h \right) \leq \prod_{s = m_I+1}^{t-1}	U_s(h,i_s). 
\end{equation}

\tcbset{colback=white}
\begin{tcolorbox}
	{\bf Third step:} Formula for $\P\left(E(h,k) = p \right)$.
\end{tcolorbox}
We can determine $\P\left(E(h,k) = p \right) = \P\left(\max_{j \in M_k}\widehat{X}_{j}(h,k) = g(p) \right)$ by using a result from \citet{Vaughan1972permanent} that describes the PDF of the maximum of random variables coming from different distributions. Note that each $\widehat{X}_{j}(h,k)$ has a different distribution\footnote{For example, if $X_i$ has a Bernoulli distribution with parameter $\mu_i$, $\widehat{X}_{i,2}$ assumes values in $\{0, 1, 2\}$, with probabilities $(1-\mu_i)^2,(1-\mu_i)\mu_i, \mu_i^2$, while $\widehat{X}_{i,3}$ assumes values in $\{0, 1, 2, 3\}$, with probabilities $(1-\mu_i)^3,(1-\mu_i)^2\mu_i, (1-\mu_i)\mu_i^2, \mu_i^3$.} that depends also on $s_j$. Given a square matrix $A$, let $\perm(A)$ be the permanent\footnote{The permanent of a square matrix $A$ is defined like the determinant, except that all signs are positive.} of $A$. Then, the PDF of $\max_{j \in M_t}\widehat{X}_{j}(h,k)$ is given by\small
\begin{equation}
f_{M(h,k)}(x)=
\frac{1}{(m_t-1)!} \perm\left(    
\begin{bmatrix}
F_{1}(x)       & F_{2}(x) &  \dots & F_{m_k}(x)   \\
\vdots         & \vdots   &  \vdots& \vdots       \\
F_{1}(x)       & F_{2}(x) &  \dots & F_{m_k}(x)   \\
f_{1}(x)       & f_{2}(x) &  \dots & f_{m_k}(x)  
\end{bmatrix}
\right)  
\begin{array}{lc}
\Bigg\}      & \vphantom{\rule{1mm}{27pt}} m_k -1 \;\text{rows}    \\
\vphantom{\rule{1mm}{17pt}}         & \vphantom{\rule{1mm}{17pt}}   
\end{array}
\end{equation}\normalsize

where $f_{1}(x), \cdots, f_{m_k}(x)$ and $F_{1}(x), \cdots, F_{m_k}(x)$ are the PDFs (or PMFs) of  the cumulative distributions of the average rewards $\widehat{X}_{j}(h,k)$ of arms $j \in M_k$ (if unknown, they are approximated by a Normal r.v. by CLT). 
Thus,
\begin{equation}
\P\left(E(h,k) = p \right) = f_{M(h,k)}(g(p)) .
\end{equation}

\tcbset{colback=white}
\begin{tcolorbox}
	{\bf Fourth step:} Formula for $\P\left(t \in I(j)\; \Big|\; H_{t-1} = h \right)$.
\end{tcolorbox}
We have that
\begin{equation}\label{Equation::I_t=j|H_t-1}
\P\left(t \in I(j)\; \Big|\; H_{t-1} = h \right) =  
\int_{0}^{1} \left[ p\frac{1}{m_t} + (1-p)\P\left(\widehat{X}_{j}(h,t) > \widehat{X}_{i}(h,t) \;\;\forall i \neq j\right) \right]f_{M(h,t)}(g(p))  \text{d}p
\end{equation}
Similarly to Step 2, $\P\left(\widehat{X}_{j}(h,t) > \widehat{X}_{i}(h,t) \;\;\forall i \neq j\right)$ has upper bound
\begin{equation}
	u_t(h,j) = 
	\prod_{i: \mu_i > \mu_{j}} \left( \exp\left\{-\frac{t_{j}(h, t) \Delta(i,j)^2}{2 r}\right\} +  \exp\left\{-\frac{t_{i}(h,t) \Delta(i,j)^2}{2 r}\right\}\right),
\end{equation}\normalsize
and \eqref{Equation::I_t=j|H_t-1} has upper bound $U_t(h,j)$, where
\begin{equation}
U_t(h,j) = \int_{0}^{1} \left( \frac{p}{m_{t}}    +
(1 - p) u_t(h,j) 
\right) f_{M(h,t)}(g(p))\;\text{d}p.
\end{equation}
Note that $U_t(h,j)$ is different from $U_s(h,i_s)$ defined in \eqref{Equation::U_s} that have values of $b_s$ available.

\tcbset{colback=white}
\begin{tcolorbox}
	{\bf Fifth step:} Bringing together all the bounds of the previous steps.
\end{tcolorbox}
From \eqref{Equation::Decomposition} we have that
\begin{equation}
	\P\left(t \in I(j)\right) = \sum_{h \in \mathcal{H}_{t-1}} \P\left(t \in I(j)\; \Big|\; H_{t-1} = h \right)\P\left( H_{t-1} = h \right)  \leq 
	\sum_{h \in \mathcal{H}_{t-1}} U_t(h,j)  \prod_{s = m_I+1}^{t-1}	U_s(h,i_s),
\end{equation}
and from \eqref{Equation::ER_Adaptive_greedy} in conclusion:
\begin{equation*}
\E[R_n]\leq \sum_{j \in M_I}\Delta_{j,i^*_{j}} +    \sum_{t=m_I + 1}^n \;\sum_{j \in M_t} \Delta_{j,i^*_{t}} \sum_{h \in \mathcal{H}_{t-1}} \left(U_t(h,j)  \prod_{s = m_I+1}^{t-1}	U_s(h,i_s)\right) .
\end{equation*}   

\newpage
\section{The regret bound of the UCB mortal algorithm}\label{Proof::UCB_mortal_regret_bound}
\tcbset{colback=blue!2!white}
\begin{tcolorbox}
	{\bf Theorem \ref{Theorem::UCB-L}}\textit{ 
	Let $\bigcup_{z=1}^{E_j}L_j^z$ be a partition of $L_j$ into epochs with different best available arm, $s_j^z$ and $l_j^z$ be the first and last step of epoch $L_j^z$, and for each epoch let $u_{j,z}$ be defined as
	\begin{equation}
	u_{j,z} = \max_{t\in\{s_j^z,\cdots,l_j^z\}}\left\lceil \frac{8 \psi(j,t) \log (t-s_j)}{\Delta_{j,z}^2}  \right\rceil,
	\end{equation}
	where
	\begin{equation}
	\Delta_{j,i^*_{t}} = \Delta_{j,z} \;\;\text{for}\;t \in L_j^z.
	\end{equation}
		Then, the bound on the mean regret $\E[R_n]$ at time $n$ is given by
		\footnotesize
		\begin{eqnarray*}
			\E[R_n] &\leq& \sum_{j \in M_I}\Delta_{j,i^*_{j}} \\ &+&   \sum_{j \in M}\; \sum_{z =1}^{E_j}  \Delta_{j,z}\min\left(l_j^z-s_j^z \;,\; u_{j,z} + \displaystyle\sum_{\substack{t \in L_j^z \\ t>m_I}}\; (t-s_{i^*_{t}}) (t-s_j-u_{j,z}+1)\left[  (t-s_j)^{-\frac{4}{r^2}\psi(j,t)} +  (t-s_{i^*_{t}})^{-\frac{4}{r^2} \psi(i^*_{t},t)}  \right] \right) .
		\end{eqnarray*}	\normalsize
	}
\end{tcolorbox}
\tcbset{colback=white}
\begin{tcolorbox}
	{\bf First step:} Decomposition of $\E[R_n]$.
\end{tcolorbox}
Let us partition the set of steps $L_j$ during which arm $j$ is available into $E_j$ epochs $L_j^z$, such that 
\begin{itemize}
	\item $\bigcup_{z=1}^{E_j}L_j^z = L_j$,
	\item $L_j^{z_1} \cap L_j^{z_2} = \emptyset$ if $z_1 \neq z_2$,
	\item $i_t^* \neq i_s^*$ if $t \in L_j^{z_1}$ and $s\in L_j^{z_2}$ (i.e., if different epochs have different best arm available).
\end{itemize}
Since during the same epoch the best arm available does not change, let us define 
\begin{equation}
	\Delta_{j,i^*_{t}} = \Delta_{j,z} \;\;\text{for}\;t \in L_j^z,
	\end{equation}
and $s_j^z = \min{L_j^z}$, $l_j^z = \max{L_j^z}$ the first and last step of epoch $L_j^z$.\\
Then, using the second formulation of the cumulative regret given in \eqref{Equation::cumulative_regret_second_form} we have that 
\begin{eqnarray}
R_n &=& \sum_{j \in M_I}\Delta_{j,i^*_{j}} +   \sum_{j \in M}\; \sum_{\substack{t \in L_j \\ t>m_I}}  \Delta_{j,i^*_{t}} \ONE{\{t \in I(j)\}} \label{Equation::R_n_to_take_expectation_0}\\
 &=& \sum_{j \in M_I}\Delta_{j,i^*_{j}} +   \sum_{j \in M}\; \sum_{z =1}^{E_j}  \Delta_{j,z} \sum_{\substack{t \in L_j^z \\ t>m_I}}  \ONE{\{t \in I(j)\}} \label{Equation::R_n_to_take_expectation}
\end{eqnarray}
Let us call
\begin{equation*}\label{Equation::T_j_total}
T_j^z(l_j^z) = \sum_{\substack{t \in L_j^z \\ t>m_I}}   \ONE{\{t \in I(j)\}}
\end{equation*}
the total number of times we choose arm $j$ in epoch $z$ during the game (after initialization). Then, by taking the expectation of \eqref{Equation::R_n_to_take_expectation} we get 
\begin{equation}\label{Equation::ER_n_UCB_mortal}
\E[R_n] = \sum_{j \in M_I}\Delta_{j,i^*_{j}} +   \sum_{j \in M}\; \sum_{z =1}^{E_j}  \Delta_{j,z}\, \E\left[ T_j^z(l_j^z) \right]. 
\end{equation}
Therefore, finding an upper bound for the expected value of \eqref{Equation::R_n_to_take_expectation_0} can be accomplished by bounding the expected value of $T_j^z(l_j^z)$.
\begin{tcolorbox}
	{\bf Second step:} Decomposition of $T_j^z(l_j^z)$.
\end{tcolorbox}
Recall that with $T_j(t-1)$ we indicate the number of times we played arm $j$ before turn $t$ starts.  For any integer $u_{j,z}$, we can write
\footnotesize
\begin{eqnarray*}
	T_j^z(l_j^z) & = &  u_{j,z} + \displaystyle\sum_{\substack{t \in L_j^z \\ t>m_I}}   \ONE{\{t \in I(j), T_j(t-1) \geq u_{j,z}\}}  \label{step_two}\\
	&= & u_{j,z} + \displaystyle\sum_{\substack{t \in L_j^z \\ t>m_I}}  \ONE\left\{ \widehat{X}_{j} + \psi(j,t)\sqrt{\frac{2 \log (t-s_j)}{T_{j}(t-1)}} > 
	\widehat{X}_{i^*_{t}} + \psi(i^*_{t},t)\sqrt{\frac{2 \log (t-s_{i^*_{t}})}{T_{i^*_{t}}(t-1)}}, T_j(t-1)\geq u_{j,z}\right\}  \label{step_three}\\
	&\leq & u_{j,z} + \displaystyle\sum_{\substack{t \in L_j^z \\ t>m_I}}\; \sum_{k_j = u_{j,z}}^{t-s_j}\; \sum_{k_{i^*_{t}} = 1}^{t-s_{i^*_{t}}} \ONE\left\{ \widehat{X}_{j} + \psi(j,t)\sqrt{\frac{2 \log (t-s_j)}{k_j}} > \widehat{X}_{i^*_{t}} + \psi(i^*_{t},t)\sqrt{\frac{2 \log (t-s_{i^*_{t}})}{k_{i^*_{t}}}}\right\}.  \label{step_four}
\end{eqnarray*}\normalsize
Therefore we can find an upper bound for the expectation of $T_j^z(l_j^z)$ by finding an upper bound for the probability of the event $$A = \left\{ \widehat{X}_{j} + \psi(j,t)\sqrt{\frac{2 \log (t-s_j)}{k_j}} > \widehat{X}_{i^*_{t}} + \psi(i^*_{t},t)\sqrt{\frac{2 \log (t-s_{i^*_{t}})}{k_{i^*_{t}}}}\right\}.$$
\begin{tcolorbox}
	{\bf Third step:} Upper bound for $\E[T_j^z(l_j^z)]$.
\end{tcolorbox}
Using Proposition \ref{Proposition::at_least_one_of_three} and Proposition \ref{Proposition::the_third_cannot_hold} we have that, by choosing $u_{j,z} = \max_{t\in\{s_j^z,\cdots,l_j^z\}}\left\lceil \frac{8 \psi(j,t) \log (t-s_j^z)}{\Delta_{j,z}^2} \right\rceil$,
\begin{equation}\label{Equation::inclusion_0}
A \subset \left(\left\{ \widehat{X}_{i^*_{t}} < \mu_{i^*_{t}} - \psi(i^*_{t},t)\sqrt{\frac{2 \log (t-s_{i^*_{t}})}{k_{i^*_{t}}}} \right\} \cup \left\{ \widehat{X}_{j} >       \mu_j         + \psi(j,t)     \sqrt{\frac{2 \log (t-s_j)}{k_j}}  \right\} \right).
\end{equation}
Using Hoeffding's\footnote{{\bf Hoeffding's bound:} Let $X_1, \cdots, X_n$ be r.v. bounded in $[a_i,b_i]$ $\forall i$. Let $\widehat{X} = \frac{1}{n}\sum_{i=1}^n X_i$ and $\mu = \E[\widehat{X}]$. \\Then, $\P\left(\widehat{X} - \mu \geq \varepsilon\right) \leq \exp\left\{ -\frac{2n^2\varepsilon^2}{\sum_{i=1}^n (b_i-a_i)^2 }  \right\}$. \\In our case, $n$ is $k_j$ or $k_{i^*_{t}}$, $b_i - a_i$ is $r$, $\mu$ is $\mu_j$ or $\mu_{i^*_{t}}$, and $\varepsilon$ is  $\psi(j,t)     \sqrt{\frac{2 \log (t-s_j)}{T_j(t-1)}}$ or $\psi(i^*_{t},t)\sqrt{\frac{2 \log (t-s_{i^*_{t}})}{T_{i^*_{t}}(t-1)}}$.} bound we have that 
\begin{eqnarray*}
	&&\P\left(\widehat{X}_{i^*_{t}} < \mu_{i^*_{t}} - \psi(i^*_{t},t)\sqrt{\frac{2 \log (t-s_{i^*_{t}})}{T_{i^*_{t}}(t-1)}}\right) \leq 
	\exp\left\{-\frac{2 k_{i^*_{t}}^2 \psi(i^*_{t},t)^2 \frac{2 \log (t-s_{i^*_{t}}) }{k_{i^*_{t}}} }{k_{i^*_{t}} r^2} \right\} =  (t-s_{i^*_{t}})^{-\frac{4}{r^2} \psi(i^*_{t},t)}\\
	&&\P\left(\widehat{X}_{j} >       \mu_j         + \psi(j,t)     \sqrt{\frac{2 \log (t-s_j)}{T_j(t-1)}}\right) \leq 
	\exp\left\{-\frac{2 k_j^2 \psi(j,t)^2 \frac{2 \log (t-s_j) }{k_j} }{k_j r^2} \right\} =  (t-s_j)^{-\frac{4}{r^2}\psi(j,t)}.
\end{eqnarray*}
Using the inclusion in \eqref{Equation::inclusion_0} in combination with Hoeffding's bounds, we have that
\begin{eqnarray}
\E\left[T_j^z(l_j^z)\right] &\leq& u_{j,z} + \displaystyle\sum_{\substack{t \in L_j^z \\ t>m_I}}\; \sum_{k_j = u_{j,z}}^{l_j}\; \sum_{k_{i^*_{t}} = 1}^{t-s_{i^*_{t}}} \P\left\{ \widehat{X}_{j} + \psi(j,t)\sqrt{\frac{2 \log (t-s_j)}{k_j}} > \widehat{X}_{i^*_{t}} + \psi(i^*_{t},t)\sqrt{\frac{2 \log (t-s_{i^*_{t}})}{k_{i^*_{t}}}}\right\}\nonumber\\
&\leq& u_{j,z} + \displaystyle\sum_{\substack{t \in L_j^z \\ t>m_I}}\; \sum_{k_j = u_{j,z}}^{t-s_j}\; \sum_{k_{i^*_{t}} = 1}^{t-s_{i^*_{t}}} \left[(t-s_{i^*_{t}})^{-\frac{4}{r^2} \psi(i^*_{t},t)} + (t-s_j)^{-\frac{4}{r^2}\psi(j,t)}\right]\nonumber\\
&=&u_{j,z} + \displaystyle\sum_{\substack{t \in L_j^z \\ t>m_I}}\; (t-s_{i^*_{t}}) (t-s_j-u_{j,z}+1)\left[  (t-s_j)^{-\frac{4}{r^2}\psi(j,t)} +  (t-s_{i^*_{t}})^{-\frac{4}{r^2} \psi(i^*_{t},t)}  \right]. \label{Equation::bound_on_ET_j_UCB_mortal_0}
\end{eqnarray}
Of course, we also have that the expected number of times the algorithm chooses arm $j$ during epoch $L_j^z$ is also bounded by the length of the epoch itself $l_j^z-s_j^z$ (this bound is useful in case the epoch is very short). Combining this with \eqref{Equation::bound_on_ET_j_UCB_mortal_0} we have that 
\begin{equation}\label{Equation::bound_on_ET_j_UCB_mortal}
	\E\left[T_j^z(l_j^z)\right] \leq  \min\left(l_j^z-s_j^z \;,\; u_{j,z} + \displaystyle\sum_{\substack{t \in L_j^z \\ t>m_I}}\; (t-s_{i^*_{t}}) (t-s_j-u_{j,z}+1)\left[  (t-s_j)^{-\frac{4}{r^2}\psi(j,t)} +  (t-s_{i^*_{t}})^{-\frac{4}{r^2} \psi(i^*_{t},t)}  \right] \right).
\end{equation}
\begin{tcolorbox}
	{\bf Fourth step:} Get upper bound for $\E[R_n]$.
\end{tcolorbox}
Combining \eqref{Equation::bound_on_ET_j_UCB_mortal} with \eqref{Equation::ER_n_UCB_mortal} we get that the bound on the cumulative regret is given by
\begin{eqnarray*}
		\E[R_n] &\leq& \sum_{j \in M_I}\Delta_{j,i^*_{j}} \\
		 &+&   \sum_{j \in M}\; \sum_{z =1}^{E_j}  \Delta_{j,z}\min\left(l_j^z-s_j^z \;,\; u_{j,z} + \displaystyle\sum_{\substack{t \in L_j^z \\ t>m_I}}\; (t-s_{i^*_{t}}) (t-s_j-u_{j,z}+1)\left[  (t-s_j)^{-\frac{4}{r^2}\psi(j,t)} +  (t-s_{i^*_{t}})^{-\frac{4}{r^2} \psi(i^*_{t},t)}  \right] \right) . 
\end{eqnarray*}

Notice that if $\psi(j,t) = 1$ , $s_j = 0$ and $l_j > n$ $\forall j, t$, you can recover the bound of the standard UCB algorithm used in the stochastic case.
(Note that you should use $P>2$ instead of $2$ when $r$ is not $1$ to create the UCB.)

\newpage
The results in Proposition \ref{Proposition::at_least_one_of_three} and \ref{Proposition::the_third_cannot_hold} are similar to arguments used in \citet[]{auer2002finite} for the proof of the regret bound for the UCB algorithm (here we have additional weighting of the upper confidence bound).
\tcbset{colback=white}
\begin{tcolorbox}
	\begin{proposition}\label{Proposition::at_least_one_of_three}
		The event
		\begin{equation*}
			A = \left\{ \widehat{X}_{j} + \psi(j,t)\sqrt{\frac{2 \log (t-s_j)}{T_j(t-1)}} > \widehat{X}_{i^*_{t}} + \psi(i^*_{t},t)\sqrt{\frac{2 \log (t-s_{i^*_{t}})}{T_{i^*_{t}}(t-1)}}\right\}
		\end{equation*}
		is included in $B \cup C \cup D$, where
		\begin{eqnarray*}
			B &=& \left\{ \widehat{X}_{i^*_{t}} < \mu_{i^*_{t}} - \psi(i^*_{t},t)\sqrt{\frac{2 \log (t-s_{i^*_{t}})}{T_{i^*_{t}}(t-1)}} \right\}\\
			C &=& \left\{ \widehat{X}_{j} >       \mu_j         + \psi(j,t)     \sqrt{\frac{2 \log (t-s_j)}{T_j(t-1)}}  \right\}\\
			D &=& \left\{ \mu_{i^*_{t}} - \mu_j < 2\psi(j,t)     \sqrt{\frac{2 \log (t-s_j)}{T_j(t-1)}} \right\}
		\end{eqnarray*}
		The inclusion $A \subset (B \cup C \cup D)$ intuitively means that if the algorithm is choosing to play suboptimal arm $j$ at turn $t$, then it is underestimating the best arm available (event $B$), or it is overestimating arm $j$ (event $C$), or it has not pulled enough times arm $j$ to distinguish its performance from the one of arm $i^*_{t}$ (event $D$).
	\end{proposition}
\end{tcolorbox}
For the sake of contradiction let us assume there exists $\omega \in A$ such that $\omega \in (B \cup C \cup D)^{\mathcal{C}}$. Then, for that $\omega$, none of the inequalities that define the events $B$, $C$, and $D$ would hold, i.e. (using, in order, the inequality in $B$, then the one in $D$, then the one in $C$):
\begin{eqnarray*}
   \widehat{X}_{i^*_{t}} &\geq& \mu_{i^*_{t}} - \psi(i^*_{t},t)\sqrt{\frac{2 \log (t-s_{i^*_{t}})}{T_{i^*_{t}}(t-1)}} \\
   &\geq& \mu_j + 2\psi(j,t)     \sqrt{\frac{2 \log (t-s_j)}{T_j(t-1)}} - \psi(i^*_{t},t)\sqrt{\frac{2 \log (t-s_{i^*_{t}})}{T_{i^*_{t}}(t-1)}} \\
   &\geq& \widehat{X}_{j} + \psi(j,t)     \sqrt{\frac{2 \log (t-s_j)}{T_j(t-1)}} - \psi(i^*_{t},t)\sqrt{\frac{2 \log (t-s_{i^*_{t}})}{T_{i^*_{t}}(t-1)}} ,
\end{eqnarray*}
which contradicts $\omega \in A$.

The result in Proposition \ref{Proposition::the_third_cannot_hold} is similar to the one used in \citet[]{auer2002finite} for the proof of the regret bound for the UCB algorithm.
\tcbset{colback=white}
\begin{tcolorbox}
	\begin{proposition}\label{Proposition::the_third_cannot_hold}
		When
		\begin{equation*}
		T_j(t-1) \geq \left\lceil \frac{8 \psi(j,t) \log (t-s_j)}{\Delta_{j,i^*_{t}}^2} \right\rceil
		\end{equation*}
		event D in Preposition \ref{Proposition::at_least_one_of_three} can not happen.
	\end{proposition}
\end{tcolorbox}
In fact, 
\begin{eqnarray*}
	&&\mu_{i^*_{t}} - \mu_j - 2\psi(j,t)     \sqrt{\frac{2 \log (t-s_j)}{T_j(t-1)}} \\
	&\geq& \mu_{i^*_{t}} - \mu_j - 2\psi(j,t)     \sqrt{\frac{2 \log (t-s_j)}{\left\lceil \frac{8 \psi(j,t) \log (t-s_j)}{\Delta_{j,i^*_{t}}^2} \right\rceil}} \\
	&\geq& \mu_{i^*_{t}} - \mu_j - 2\psi(j,t)     \sqrt{\frac{ \log (t-s_j) \Delta_{j,i^*_{t}}^2}{ 4 \psi(j,t) \log (t-s_j) }} \\
	&=& \mu_{i^*_{t}} - \mu_j - \Delta_{j,i^*_{t}} = 0.
\end{eqnarray*}

\normalsize

\newpage

\section{Useful results}
The result in Proposition \ref{Proposition::inclusion} is similar to the one used in \citet[]{auer2002finite} for the proof of the regret bound for the $\varepsilon$-greedy algorithm.
\tcbset{colback=white}
\begin{tcolorbox}
	\begin{proposition}\label{Proposition::inclusion}
		Let $\mu_i > \mu_j$ and let us define the following events:\small
		\begin{eqnarray*}
			A &=& \left\{ \widehat{X}_{j} > \widehat{X}_{i} \right\},\\
			B &=& \left\{ \widehat{X}_{i}  <  \mu_i - \frac{\Delta(i,j)}{2} \right\},\\
			C &=& \left\{ \widehat{X}_{j} > \mu_j + \frac{\Delta(i,j)}{2}\right\}.
		\end{eqnarray*}
		\normalsize Then,
		\begin{equation}\label{__inclusion1}
		A \subset \left(  B \cup  C  \right).
		\end{equation}
	\end{proposition}
\end{tcolorbox}
Intuitively, the inclusion in \eqref{__inclusion1} means that we play arm $j$ when we underestimate the mean reward of the best arm, or when we overestimate that of arm $j$.
Assume for the sake of contradiction that there exists an element $\omega \in A$ that does not belong to $B \cup C$. Then, we have that $\omega \in \left(B \cup C\right)^C$
\begin{eqnarray}
\Rightarrow \;\;\omega & \in & \left(  \left\{ \widehat{X}_{i}  <  \mu_i - \frac{\Delta(i,j)}{2}  \right\} \cup 
\left\{ \widehat{X}_{j} > \mu_j + \frac{\Delta(i,j)}{2} \right\}  \right)^C \label{__toContradict1} \\
\Rightarrow \;\;\omega &\in&   \left\{ \widehat{X}_{i}  \geq  \mu_i - \frac{\Delta(i,j)}{2}  \right\} \cap 
\left\{ \widehat{X}_{j} \leq \mu_j + \frac{\Delta(i,j)}{2}  \right\}. \label{__NONinclusion1}
\end{eqnarray}
By definition we have $\mu_i - \frac{\Delta(i,j)}{2} = \mu_i - \frac{\mu_i-\mu_j}{2} = \frac{\mu_i+\mu_j}{2} =  \mu_j + \frac{\Delta(i,j)}{2}$. From the inequalities given in \eqref{__NONinclusion1} it follows that  
\begin{eqnarray*}
	\widehat{X}_{i} \geq \mu_i - \frac{\Delta(i,j)}{2} = \mu_j + \frac{\Delta(i,j)}{2}  
	\geq \widehat{X}_{j},
\end{eqnarray*}
but this contradicts our assumption that $ \omega \in A = \left\{ \widehat{X}_{j}  > \widehat{X}_{i}  \right\}$.\\ 
Therefore, all elements of $A$ belong to $B \cup C$.

\newpage
\section{Numerical results}
\subsection{Dataset}
The dataset can be found on the Yahoo Webscope program. 
It contains files recording 15 days of article recommendation history. Each record shows information about the displayed article id, user features, timestamp and the candidate pool of available articles at that time. The displayed article id shows the arm that recommenders pick each turn. User features were not used, since our algorithms look for articles generally liked by everyone. Timestamp tells the time that an event happens; along with the candidate pool of available articles, we can scan through the records and find out each article's lifespan.

\subsection{Evaluation methodology}
A unique property of this dataset is that the displayed article is chosen
uniformly at random from the candidate article pool. Therefore, one can use an unbiased offline evaluation method \cite{Li:2011:UOE:1935826.1935878} to compare bandit algorithms in a reliable way. However, in the initialization phase, we applied a simpler and faster method (Algorithm \ref{Initialization}), since initialization only plays 25 turns in a game and we care more about what happens later on.

In order to apply these evaluation methods, after parsing the original text log into structured data frame, we made an event stream generator out of it. The event stream generator has a member method ``next\_event()'' that gives us the next record in the data frame. The fields in the record give information about the event. For example, in the initialization phase we checked the ``article'' field of the records to see if that article had been played before.

\begin{algorithm}
	\caption{Initialization}
	\begin{algorithmic}\label{Initialization}
		\STATE event stream $ Stream $
		\STATE number of turns as initialization $ m $
		\STATE $ i \gets 0 $
		\WHILE{$ i<m $}
		\STATE $ Record \gets Stream.next\_event() $
		\IF{$Record.article$ was not seen before}
		\STATE update expectation of $Record.article$
		\STATE $ i \gets i+1 $
		\ENDIF
		\ENDWHILE
	\end{algorithmic}
\end{algorithm}

\subsection{Parameter tuning}
AG-L filters out a portion of articles that expire soon. This portion is a tunable parameter. We tested different values with a smaller size dataset and finally used 0.1 as the threshold. In UCB-L's upper confidence bound, $ \psi(j,t) = c \log (l_j - t + 1 ) $ and $ c $ is a tunable parameter. After tuning, we set $ c=0.011 $ for later experiments.

\subsection{UCB score function}
The original expression for the modified upper confidence bound in UCB-L is $ \hat{X}_{j} + \psi(j,t)\sqrt{\frac{2 \log (t-s_j)}{T_{j}(t-1)}} $. In the experiment, we used $ \hat{X}_{j} + \psi(j,t)\sqrt{\frac{2 \log (t-s_j +1)}{T_{j}(t-1)}} $ to avoid an invalid value when an article is chosen the turn it becomes available  ($ t=s_j $).

\subsection{Timestamp vs Turn number}

In this offline evaluation setting, a considerable portions of events are discarded if they do not match the actions that are chosen by our algorithms. Each event has a timestamp, but there is no direct relation between an event's timestamp and a turn in the bandit game (we denote a generic turn number with $t$). Since timestamps and turn numbers are positively correlated, we can use the set of timestamps as a proxy to rank articles by remaining life. Given the rank of remaining lifespan, AG-L plays only the arms at the top of the rank. With this proxy, we are able to simulate the AG-L algorithm pretending we know the exact lifespan of an article (in addition to the case where we estimate the lifespans of the articles). 

For UCB-L however, the ranking of the arms is not sufficient. UCB-L needs to know the exact turns $s_j$ at which an article $j$ is available or turn $l_j$ at which it stops to be available. Since we can not map timestamps to turns, we only simulated the case of UCB-L estimating the life of articles.



At the beginning of the game, we can not estimate correctly lifespans because we have not yet seen an expired article. If  our estimated life length $ \hat{L} $ is too small, then it can happen that $ \hat{L} +s_j-t+1 \leq 0 $ , yielding an invalid value for $ \psi(j,t) = c \log(l_j - t + 1) = c \log( \hat{L} +s_j-t+1) $. In these cases we set $ l_j - t + 1 = \hat{L} +s_j-t+1 = 0 $ and use only $ \hat{X}_j $ as the upper confidence bound.

\subsection{Contextual algorithm}
Algorithm \ref{Algorithm::LinUCB-L} is a similar adaptation of the LinUCB algorithm introduced by \citet{li2010contextual} to the mortal setting. Also in this case, the function $\psi(j,t)$ regulates the amplitude of the upper confidence bound above the estimated mean according to the remaining life of the arm.
As before, new arms are initialized by using the average performance of past arms (i.e., if in the past a lot of bad arms appeared, new arms are considered more likely to be bad, and vice-versa if lots of good arms appeared in the past).
\RestyleAlgo{boxruled}
\begin{algorithm}[]
	\caption{LinUCB-L algorithm}\nllabel{Algorithm::LinUCB-L}
	\SetKwInOut{Input}{Input}
	\SetKwInOut{Output}{output}
	\SetKwInOut{Loop}{Loop}
	\SetKwInOut{Initialization}{Initialization}
	\Input{number of rounds $n$, initial set of arms $M_I$, set $M_t$ of available arms at time,  rewards range $[a,b]$, dimension $d$ (context space dimension $+$ arms space dimension)}
	\Initialization{For each $j \in M_I$, $A_j = I_d$, $b_j = 0_{d\times 1}$}
	\For{$t=1$ \KwTo $n$}{
		{Get context $x_t$ (or $x_{t,j}$ if each arm gets its context)\;}
		\For{$j=1$ \KwTo $m_t$}{
			{Set $\hat{\theta}_j = A^{-1}_j b_j$\;}
			{Set $UCB_j = \hat{\theta}_j^{T}x_t + \psi(j,t)\sqrt{x_t^TA^{-1}_j x_t}$\;}
		}
		{Play arm $j=\argmax_i UCB_i$\;}
		{Get reward $X_j(t)$\;}
		{Update $A_j = A^{-1}_j + x_tx_t^{T}$\;}
		{Update $b_j = b_j + r_tx_t$\;}
	}
\end{algorithm}

We have noticed that the contextual algorithm was not useful for the features made available in the Yahoo! Webscope Dataset, so for the experiments we used the non-contextual version presented in the main paper.

\newpage
\small
\section{Notation summary}\nllabel{Section::notation}
\tcbset{colback=gray!5!white}
\begin{tcolorbox}
	\begin{itemize}
		\item $M_t$ as the set of all available arms at turn $t$;	
		\item $M_I$ the set of arms that are initialized;
		\item $m_t$: number of arms available at time $t$;
		\item $n$: total number of rounds;
		\item $X_j(t)$: random reward for playing arm $j$ at time $t$;
		\item $\mu_{*}$: mean reward of the optimal arm ($\mu_{*} = \max_{1\leq j \leq m} \mu_j$);
		\item $\Delta(i,j)$: difference between the mean reward of arm $i$ and arm $j$  ($\Delta(i,j)=\mu_i-\mu_j$);
		\item $\hat{X_j}$: current estimate of $\mu_j$;
		\item $I_j$: set of turns when arm $j$ is played; 
		\item $T_j(t-1)$: r.v. of the number of times arm $j$ has been played before round $t$ starts;
		\item $\mathcal{H}_{t-1}$: set of all possible histories $h$ (after deterministic initialization) of the game up to turn $t-1$;
		\item $U_s(h,i_s)$: upper bound on the probability that arm $i_s$ was pulled at time $s$ given the history of pulls $h$ up to time $s-1$;
		\item $u_s(h,i_s)$: upper bound on the probability that arm $i_s$ is considered to be the best arm at time $s$ given the history of pulls $h$ up to time $s-1$;
		\item $f_{M(h,s)}(g(p))$: the PDF (or PMF) of the maximum of the estimated mean rewards at time $s$ given that each arm has been pulled according to history $h$ up to time $s-1$;
		\item $g(p)$: linear transformation $g(p) = b+(a-b)p$;
		\item $U_t(h,j)$: upper bound on the probability that arm $j$ was pulled at time $t$ given the history of pulls $h$ up to time $t-1$;
		\item $u_t(h,j)$: upper bound on the probability that arm $j$ is considered to be the best arm at time $t$ given the history of pulls $h$ up to time $t-1$;
		\item $R_n$: total regret at round $n$.
	\end{itemize}
\end{tcolorbox}
\normalsize

\end{document}